\documentclass[runningheads]{llncs}

\usepackage[T1]{fontenc}
\usepackage{graphicx}
\usepackage{amsmath,amssymb,amsfonts}
\usepackage{booktabs}
\usepackage{hyperref}
\usepackage{url}
\usepackage{microtype}
\usepackage{nicefrac}
\usepackage{pifont}
\usepackage[vlined,ruled,linesnumbered,noend]{algorithm2e} % Import algorithm2e package
\usepackage[capitalize,noabbrev]{cleveref}
\usepackage{multirow}
\usepackage{siunitx}
\usepackage{wrapfig}

\newcommand{\CMethod}[1]{\textsc{#1}}

\newcommand{\iborf}{\CMethod{i-borf}}
\newcommand{\borf}{\CMethod{borf}}
\newcommand{\CDatasetAbbr}[1]{\texttt{{#1}}}

\usepackage{xcolor}
\newcommand{\rev}[1]{{#1}}

\begin{document}

\title{A Time-Aware Bag-of-Receptive-Fields for Interpretable Irregular Time Series Classification}

\titlerunning{A Time-Aware Bag-of-Receptive-Fields}

\author{Francesco Spinnato\inst{1,2}\orcidID{0000-0002-3203-6716}
        }

\authorrunning{F. Spinnato}

\institute{Department of Computer Science, University of Pisa, Italy\\
           \email{francesco.spinnato@unipi.it}
           \and
           ISTI-CNR, Pisa, Italy}

\maketitle

% -----------------------------------------------------------------------
\begin{abstract}
Irregular time series, characterized by non-uniform sampling intervals, missing observations, and variable lengths, are ubiquitous in healthcare, mobility, and environmental monitoring, yet effective and interpretable classifiers for this setting are limited. Existing approaches often rely on imputation, which can obscure the temporal structure of the data, or require complex neural architectures that are opaque and difficult to explain. In this work, we extend the Bag-Of-Receptive-Fields (BORF), a fast, deterministic, and interpretable transform for time series, to the irregular setting. Our key contribution is a time-weighted normalization scheme in which each observation is weighted proportionally to its associated time delta, making pattern extraction sensitive to the actual temporal distribution of samples rather than only their index position. This requires deriving an efficient sliding-window recurrence for the time-weighted standard deviation, preserving the linear time complexity of BORF. We benchmark the resulting method against state-of-the-art irregular time series classifiers on datasets from the \texttt{pyrregular} repository, demonstrating competitive classification performance with the added benefit of human-interpretable explanations.

\end{abstract}

\section{Introduction}
\label{sec:introduction}

Temporal data are central to domains such as healthcare, mobility analytics, environmental monitoring, and industrial sensing~\cite{shumway2000time}. In practice, however, such data are rarely collected on a perfectly regular grid. Sensors may operate at different frequencies, measurements may cover different time spans, and failures or acquisition policies may produce gaps and missing values~\cite{harvey1998messy}. The resulting irregular time series exhibit non-uniform sampling intervals, partial observation, and variable lengths. These properties affect the meaning of local temporal patterns and the validity of explanations produced by a classifier.

Time series classification (TSC) is well established for regularly sampled data, with mature repositories and large empirical comparisons across model families~\cite{bagnall2017great,middlehurst2024bakeredux}. The irregular setting is less standardized: many studies rely on narrow benchmarks or artificially remove observations from regular series~\cite{weerakody2021review,mitra2023learning}. A recent benchmark~\cite{spinnato2026pyrregular} addresses this gap by evaluating irregular TSC methods under a unified representation and experimental protocol. Its results suggest that simple generalist methods originally designed for regular data, including \CMethod{rocket}~\cite{dempster2020rocket} and \borf{}~\cite{spinnato2024fast}, can be competitive with purpose-built irregular classifiers, even when timestamps are not explicitly exploited.

This motivates revisiting symbolic and dictionary-based methods for irregular TSC. Such methods represent a time series through counts of local symbolic patterns rather than through latent neural states or randomized feature maps~\cite{baydogan2013bag,lin2007experiencing}. The Bag-Of-Receptive-Fields transform, \borf{}, follows this line by constructing a deterministic sparse representation from symbolic descriptions of local receptive fields~\cite{spinnato2024fast}. Its features are explicit, countable, and associated with concrete local patterns, making the representation naturally more inspectable than many black-box alternatives. The original \borf{} formulation, however, is designed for regularly sampled data. Its local summaries and normalizations are based on observation order, implicitly assuming that consecutive observations have equal temporal support. This assumption breaks down under irregular sampling. A dense burst of observations over a short interval can dominate local statistics, while a sparse region spanning a longer interval can be underrepresented. Uniform local averaging therefore summarizes observations rather than elapsed time, discarding information carried by the sampling structure itself.

We present \iborf{} (Irregular Bag-Of-Receptive-Fields), an extension of \borf{} for irregular time series. The central idea is to preserve the symbolic, deterministic, and interpretable structure of \borf{}, while replacing its index-based local statistics with time-weighted aggregation and normalization. This allows the symbolic representation to reflect temporal support rather than only observation order, without first imputing missing values or resampling the data onto a regular grid. To make this efficient, we derive the sliding-window weighted variance update required by the transform. Online weighted variance updates and removal by negative weights have been studied previously~\cite{west1979updating}, and Welford-style estimators are classical for expanding windows~\cite{welford1962note}; here, we specialize the recurrence to moving receptive fields with one incoming and one outgoing weighted observation.

Our contributions are as follows. \textbf{(1)} We extend SAX and the Bag-Of-Receptive-Fields transform to irregular time series through time-weighted aggregation and normalization. \textbf{(2)} We derive the sliding-window weighted variance update used by \iborf{}, enabling efficient normalization over moving receptive fields with non-uniform temporal supports. \textbf{(3)} We evaluate \iborf{} on irregular TSC datasets, comparing it against state-of-the-art irregular classifiers. \textbf{(4)} We present explainability examples for irregular TSC by tracing influential symbolic features back to timestamped temporal regions.

The remainder of this paper is organized as follows.
\Cref{sec:related} surveys related work.
\Cref{sec:background} recaps the \borf{} framework and irregular TSC setting.
\Cref{sec:method} presents \iborf{}.
\Cref{sec:experiments} reports experiments.
\Cref{sec:conclusions} concludes.

% -----------------------------------------------------------------------
\section{Related Work}
\label{sec:related}

\textbf{Irregular Time Series Classification.}
Irregular time series classification has been addressed by several specialised modelling paradigms~\cite{wang2024deep}. These include continuous-time models based on neural ODEs and CDEs~\cite{kidger2020neural}, recurrent architectures with time-decay or missingness mechanisms~\cite{che2018recurrent,cao2018brits}, graph-based models for inter-sensor dependencies under missingness~\cite{zhang2021graph}, and attention-based models with continuous-time positional encodings or imputation-oriented self-attention~\cite{shukla2021multitime,du2023saits}. While effective for specific irregularity patterns, these methods often require substantial model selection and remain difficult to interpret~\cite{spinnato2026pyrregular}.
In contrast, recent TSC bake-offs have highlighted the strength of generalist models, designed to perform robustly across heterogeneous datasets with limited task-specific tuning~\cite{middlehurst2024bakeredux}.
Kernel-based transforms such as \CMethod{rocket} exemplify this trend, combining strong empirical performance with simple downstream classifiers~\cite{dempster2020rocket}.
The Bag-of-Receptive-Fields (\CMethod{borf}) follows the same generalist philosophy, replacing random convolutional features with a deterministic dictionary-based representation built from local receptive fields~\cite{spinnato2024fast}.
A recent benchmark on irregular TSC~\cite{spinnato2026pyrregular} showed that these models remain surprisingly competitive even when originally designed for regular data, with \CMethod{rocket} emerging as the strongest overall method and \CMethod{borf} among the best-performing alternatives.
This work builds on that evidence by extending \CMethod{borf} to ITS, preserving its generalist and interpretable character while incorporating time-aware operations suited to irregular sampling.

\medskip
\noindent
\textbf{Dictionary-Based Time Series Classification.}
Dictionary-based methods discretize time series into symbolic words and build a bag-of-patterns representation~\cite{baydogan2013bag}. The Symbolic Aggregate approXimation (SAX)~\cite{lin2007experiencing} and its window-wise extension underpin many such approaches, including \CMethod{bop}~\cite{lin2012rotation}, \CMethod{mr-seql}~\cite{le2019interpretable}, and \borf{}~\cite{spinnato2024fast}.
Because these methods extract features over many overlapping windows, their scalability depends on efficient updates of local statistics, from fast Fourier updates in spectral dictionaries to stable online mean and variance estimators such as Welford's method and weighted extensions thereof~\cite{welford1962note,west1979updating}. 
All of these methods assume regularly-sampled, fixed-length inputs; to the best of our knowledge, our work is the first to extend the SAX-based dictionary paradigm to ITS by replacing uniform aggregation with time-weighted aggregation.

\medskip
\noindent
\textbf{Explainability for Time Series.}
Explainability for time series is a vast research area~\cite{theissler2022explainable}. We focus on the line most directly connected to \borf{}: SHAP-based attribution over interpretable symbolic representations. While SHAP~\cite{lundberg2017unified} can be applied to any TSC model, explanations over raw timestamps or learned representations are often computationally costly and difficult to relate to recurring temporal patterns. \borf{}~\cite{spinnato2024fast} addresses this by combining pattern frequency counts with a linear classifier, yielding local saliency maps and global pattern importance scores. For ITS, comparable explanation mechanisms remain largely unexplored, since irregular sampling and missing observations complicate temporal attribution. \iborf{} inherits the interpretable feature space of \borf{} and extends its SHAP-based explanation framework to irregular time series.

% -----------------------------------------------------------------------
\section{Background}
\label{sec:background}

In this section, we present all the necessary concepts to understand our proposal. We begin by defining an irregular time series data.

\begin{definition}[Irregular Time Series Signal]
  An \emph{irregular signal} is a sequence of $m$ observations each paired
  with a timestamp,
  $\mathbf{x} = [(x_1, t_1), \ldots, (x_m, t_m)]$,
  where $t_1 < \cdots < t_m$ and the intervals $t_{i+1} - t_i$ are not
  necessarily constant.
\end{definition}

\begin{definition}[Irregular Time Series]
  An irregular time series is a collection of $c$ potentially irregular time series signals,
  $\mathbf{X} = \{\mathbf{x}_1, \ldots, \mathbf{x}_c\}$.
\end{definition}

\begin{definition}[Irregular Time Series Dataset]
  An \emph{irregular time series dataset}
  $\mathcal{X} = \{\mathbf{X}_1, \ldots, \mathbf{X}_n\}$
  is a collection of $n$ irregular time series.
\end{definition}

\noindent Irregularity can manifest in different ways, such as \emph{uneven sampling},
\emph{partial observation} (missing values), and \emph{raggedness} (sampling, length
or alignment mismatches)~\cite{spinnato2026pyrregular}, which can affect downstream tasks, in our setting, classification. 

\begin{definition}[Irregular Time Series Classification]
  Given an irregular dataset $\mathcal{X}$,
  \emph{Irregular TSC} is the task of training a
  model $f$ to predict a categorical label $y$ for each input time series
  $\mathbf{X}$, i.e.,
  $f(\mathcal{X}) = [f(\mathbf{X}_1), \ldots, f(\mathbf{X}_n)]
   = \mathbf{\hat{y}} \in \mathbb{N}^n$.
\end{definition}

\subsection{Bag-Of-Receptive-Fields}
\label{sec:borf_recap}

\begin{wrapfigure}{t}{0.45\textwidth}
    \centering
    \includegraphics[width=1\linewidth]{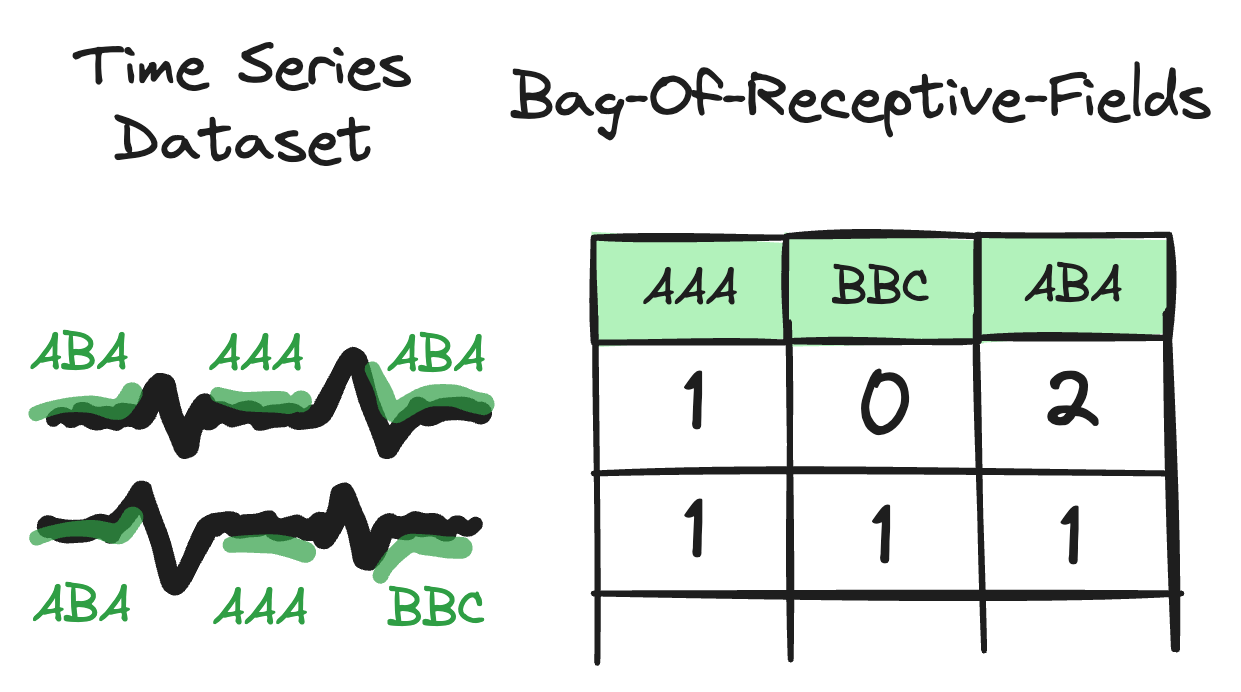}
    \caption{Bag-of-Receptive-Fields toy representation for two time series.}
    \label{fig:borf}
\end{wrapfigure}

\borf{}~\cite{spinnato2024fast} is a deterministic and interpretable transform for regular time series classification. It extends the Bag-Of-Patterns representation~\cite{baydogan2013bag} by replacing contiguous sliding windows with \emph{receptive fields}, and by combining this more flexible pattern extraction with the Symbolic Aggregate approXimation (SAX) pipeline~\cite{lin2007experiencing}. As in Bag-of-Words models for text, the final representation does not store the raw subsequences themselves, but the frequency with which symbolic patterns occur in each signal, as shown in \Cref{fig:borf}.

Let $\mathbf{x}=[x_1,\ldots,x_m]$ be a univariate regularly sampled signal. For a window size $w$, dilation $d$, and stride $s$, the $i$-th receptive field is the ordered subsequence $[x_i,x_{i+d},\ldots,x_{i+d(w-1)}]$.
The dilation controls the spacing between consecutive observations inside the receptive field, allowing the method to capture patterns at different temporal resolutions, while the stride controls the distance between consecutive receptive fields. 

\begin{figure}[t]
\centering
    \includegraphics[trim = 0mm 0mm 0mm 0mm, clip, width=0.49\textwidth]{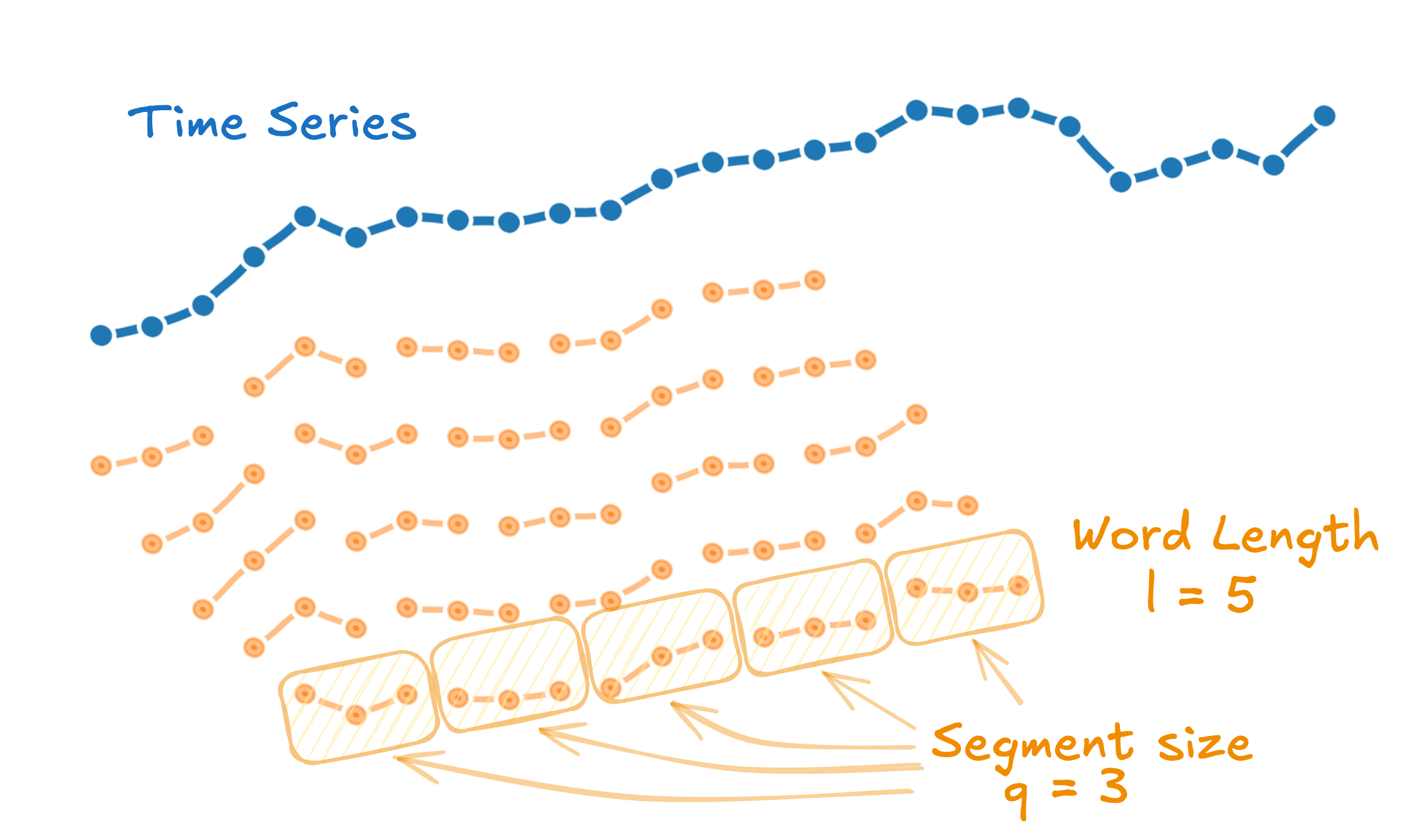}
    \includegraphics[trim = 0mm 0mm 0mm 0mm, clip, width=0.49\textwidth]{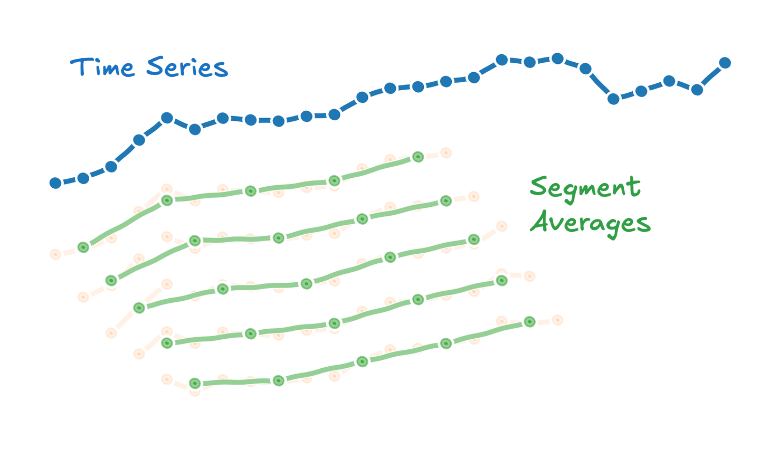}
    \caption{Naive PAA diagram used in the classical Bag-of-Patterns. The time series is divided into sliding windows, each window is segmented, and the average is taken.}
    \label{fig:paanaive}
\end{figure}

Each receptive field is then converted into a SAX word. First, the receptive field is partitioned into $l$ non-overlapping segments of equal size $q=w/l$. Second, Piecewise Aggregate Approximation (PAA) is applied by computing the mean of each segment. This produces, for every receptive field, a vector of $l$ segment means (\Cref{fig:paanaive}). Third, segment means are normalized with respect to the mean and standard deviation of the enclosing receptive field. For the $i$-th receptive field and its $j$-th segment mean $\mu_{i,j}$: 
$$
\mu^*_{i,j} = \frac{\mu_{i,j}-\mu^{\mathrm{win}}_i}{\sigma^{\mathrm{win}}_i},
$$
where $\mu^{\mathrm{win}}_i$ and $\sigma^{\mathrm{win}}_i$ are the mean and standard deviation of the full receptive field. Finally, the normalized values are discretized using Gaussian breakpoints into an alphabet of size $\alpha$, yielding a symbolic word of length $l$. Thus, each receptive field is mapped from a numerical subsequence to a SAX word. After discretization, each SAX word is hashed and identical words are counted. The final representation is a sparse contains for each time series the frequency of appearance of each symbolic receptive field (\Cref{fig:borf}).

Because each feature corresponds to the count of a specific symbolic pattern in a specific signal, the representation remains interpretable. In the original \borf{} pipeline, a linear classifier such as ridge regression can be trained on the resulting sparse representation, and feature attributions can be mapped back to symbolic words and their temporal locations. This makes it possible to identify which patterns contribute to a prediction and where they occur in the original signal.

A central contribution of \borf{} is the efficient computation of window-wise SAX words. Instead of recomputing each segment mean independently, which can be quadratic in the signal length, \borf{} precomputes moving averages for the segment size $q$. PAA segment means, window means, and window standard deviations are then obtained by lookup, so the cost depends only on the number of receptive fields and the word length.

The limitation is that SAX normalization assumes regularly sampled observations: receptive fields are indexed by position, segment means are unweighted, and window statistics ignore timestamps. 
We target this step specifically: the \borf{} pipeline is retained, while regular SAX normalization is replaced with a time-aware normalization for irregular time series.

\section{Irregular Bag-Of-Receptive-Fields}
\label{sec:method}

We introduce \iborf{} (Irregular Bag-Of-Receptive-Fields), an extension of \borf{} for irregular time series. 
The method modifies \rev{the PAA and normalization stages} of the original pipeline by replacing uniform means and standard deviations with time-weighted counterparts, computed efficiently through a sliding-window recurrence for time-weighted statistics. All subsequent steps, namely receptive-field extraction, symbolic discretization, hashing, and counting, follow the \borf{} pipeline described in \Cref{sec:background} and are therefore not repeated here.

\subsection{Time-Delta Weights}
\label{sec:weights}

When observations are non-uniformly spaced, a plain average over a receptive field treats a densely-sampled interval and a sparsely-sampled one identically, ignoring the information carried by the timestamps.
A densely-sampled observation covers only a brief moment in time, while a
sparsely-sampled one represents a longer interval and should contribute
proportionally more to any temporal aggregate.
We assign to each observation $x_i$ a weight equal to the time elapsed
since the previous observation:
\begin{equation}
  \delta_i = t_i - t_{i-1}, \quad i > 1.
  \label{eq:weights}
\end{equation}
Intuitively, $\delta_i$ approximates the length of the time interval
represented by $x_i$, making the weighted mean a discretization of a time
integral over the signal.
\Cref{eq:weights} is undefined for $i = 1$, as the first observation has
no predecessor. Several boundary conditions are reasonable in practice. \rev{In our implementation, we set $\delta_1=\delta_2$, assigning the first observation the same temporal support as the second. Investigating the sensitivity to alternative boundary conventions is left for future work.}
When all $\delta_i$ are equal, the weights cancel and every time-weighted
quantity defined below reduces to its uniform counterpart in \borf{}.

\subsection{Time-Weighted Piecewise Aggregate Approximation}
\label{sec:wpaa}

Let $\mathbf{x}$ be an irregular signal with weights $[\delta_1, \ldots, \delta_m]$
defined by \Cref{eq:weights}.
For the receptive field starting at index $i$, with window size
$w$, dilation $d$, stride $s$, word length $l$, and segment size $q = w/l$, let $a_{ijk} = 1 + (i-1)s + (j-1)dq + (k-1)d$ denote the signal index of the $k$-th element of segment $j$ in window
$i$.
The \emph{time-weighted segment mean} for segment $j$ is then:
\begin{equation}
  \hat{\mu}_{i,j} =
  \frac{\displaystyle\sum_{k=1}^{q} \delta_{a_{ijk}}\, x_{a_{ijk}}}
       {\displaystyle\sum_{k=1}^{q} \delta_{a_{ijk}}}.
  \label{eq:wpaa_weighted}
\end{equation}
This is a time-aware generalization of the PAA mean: if all $\delta_i$
are equal, it reduces to the original uniform mean of \borf{}.

\subsection{Time-Weighted Standardization}
\label{sec:wnorm}

Each segment mean $\hat{\mu}_{i,j}$ is standardized by the weighted mean
and standard deviation of its enclosing window:
\begin{equation}
  \hat{\mu}_{i,j}^* =
  \frac{\hat{\mu}_{i,j} - \hat{\mu}_{i}^{\mathrm{win}}}
       {\hat{\sigma}_{i}^{\mathrm{win}}},
  \label{eq:zscore_weighted}
\end{equation}
where $\hat{\mu}_{i}^{\mathrm{win}}$ and $\hat{\sigma}_{i}^{\mathrm{win}}$
are the weighted mean and standard deviation over all $w$ observations in
window $i$, with possibly dilated indices $i, i+d, \ldots, i+d(w-1)$. Let $W_{i}^{\mathrm{win}} = \sum_{k=0}^{w-1} \delta_{i+kd}$ be the sum of the window weights, then:

\begin{align}
  \hat{\mu}_{i}^{\mathrm{win}} &=
    \frac{1}{W_{i}^{\mathrm{win}}}
    \sum_{k=0}^{w-1} \delta_{i+kd}\, x_{i+kd},
  \label{eq:wmean_win} \\
  (\hat{\sigma}_{i}^{\mathrm{win}})^2 &=
    \frac{1}{W_{i}^{\mathrm{win}}}
    \sum_{k=0}^{w-1} \delta_{i+kd}
    \bigl(x_{i+kd} - \hat{\mu}_{i}^{\mathrm{win}}\bigr)^2.
  \label{eq:wstd_win}
\end{align}

\noindent In a regular signal all $\delta_i$ are equal and
\Cref{eq:zscore_weighted,eq:wmean_win,eq:wstd_win} reduce exactly
to the uniform z-score of the original \borf{}. 
\rev{In the irregular case, the weighting reduces the influence of densely sampled observations within each receptive field, with receptive-field boundaries, dilation, stride, and counts remaining index-based.}

\subsection{Sliding-Window Weighted Standard Deviation}
\label{sec:recurrence}
The main computational issue introduced by time-weighted normalization is the repeated computation of weighted means and standard deviations over overlapping windows. Computing these quantities independently for each of the windows takes $O(m^2)$ in the worst case.
A recurrence that updates the weighted mean and variance in $O(1)$ per window slide reduces the total cost to $O(m)$; Welford-style algorithms~\cite{welford1962note} achieve this while remaining numerically stable, avoiding the catastrophic cancellation that plagues the naïve two-pass formula in an online setting~\cite{welford1962note}.
We therefore derive a sliding-window weighted variance recurrence. Conceptually, the recurrence extends online variance updates in the spirit of~\cite{west1979updating}, but differs from them because the window both receives a new observation and removes an old one at every step.
Let $W_i = \sum_{r=i}^{i+w-1} \delta_r$ be the total weight of the
window $[x_i, \ldots, x_{i+w-1}]$, 
Sliding the window by one step removes $x_i$ (weight $\delta_i$) and
adds $x_{i+w}$ (weight $\delta_{i+w}$).
The weighted mean update follows directly from the definition:
\begin{equation}
  \hat{\mu}_{i+1} =
  \frac{W_i\,\hat{\mu}_i
        - \delta_i\, x_i
        + \delta_{i+w}\, x_{i+w}}
       {W_{i+1}}.
  \label{eq:wmean_update}
\end{equation}

\noindent Define now the weighted sum of squared deviations from the mean:
\begin{equation}
  S_i = \sum_{r=i}^{i+w-1} \delta_r\,(x_r - \hat{\mu}_i)^2
       = W_i\cdot(\hat{\sigma}_i)^2,
  \label{eq:S_def}
\end{equation}
the variance update is the following:

\begin{proposition}[Sliding-window weighted variance update]
  \label{prop:wvar}
  \begin{align}
    S_{i+1} = S_i
      &+ \underbrace{
           \delta_{i+w}(x_{i+w} - \hat{\mu}_i)
                       (x_{i+w} - \hat{\mu}_{i+1})
         }_{\text{incoming}} - \underbrace{
           \delta_i\,(x_i - \hat{\mu}_i)
                     (x_i - \hat{\mu}_{i+1})
         }_{\text{outgoing}},
    \label{eq:wvar_update}
  \end{align}
  and thus $\hat{\sigma}_{i+1} = \sqrt{S_{i+1} / W_{i+1}}$.
\end{proposition}

\noindent 
The proof is presented in Appendix \ref{sec:derivation}.
This procedure assumes dilation $d=1$; dilation $d>1$ can be handled by applying the recurrences independently to each of the $d$ interleaved sub-sequences $[x_{r}, x_{r+d}, x_{r+2d}, \ldots]$ for $r \in \{1, \ldots, d\}$. 
Computing all weighted moving averages and sliding-window standard
deviations via \Cref{eq:wmean_update,prop:wvar} requires $O(m)$ per
sub-sequence, hence $O(m)$ overall for any fixed $d$, identical to the original \borf{}.
\rev{Indeed, the additional weighted updates require constant time per window shift and therefore do not change the $O(m)$ asymptotic complexity.}

% -----------------------------------------------------------------------
\section{Experiments}
\label{sec:experiments}

\medskip
\noindent
\textbf{Datasets.}
We evaluate our proposal on $15$ datasets provided by the \texttt{pyrregular} framework~\cite{spinnato2026pyrregular}, covering health, mobility, human activity recognition, and sensor domains, \rev{using its default preprocessing and settings.}
We select all the datasets that exhibit missing values, uneven sampling, or ragged sampling, since these irregularity types directly affect the computation of local statistics and can therefore benefit from time-aware normalization. In contrast, datasets characterized only by unequal length or temporal shift are excluded, as these irregularities do not by themselves require modifying the SAX normalization step. We use the default train/test splits and report macro-averaged F1, which is robust to class imbalance. Code is available at \url{https://github.com/fspinna/borf}. 

\medskip
\noindent
\textbf{Competitors.}
We benchmark \iborf{} against reference results of state-of-the-art models from~\cite{spinnato2026pyrregular}. 
\iborf{} uses the same heuristic as \borf{}~\cite{spinnato2024fast}:
$w \in [2^2, \ldots, 2^{\lfloor\log_2 m\rfloor}]$,
dilations $d \in [2^0, \ldots, 2^{\lfloor\log_2\log_2 m\rfloor}]$,
word lengths $l \in \{2, 4, 8\}$, alphabet size $\alpha = 3$,
stride $s = 1$, and threshold $\theta = 0.15$. However, differently from the classical \borf{} implementation, \iborf{} uses Ridge instead of the classical \CMethod{lgbm}. Thus, to isolate the effect of the proposed weighting scheme, we also include the classical \borf{} with a Ridge head (\CMethod{borf\texttt{+}rc}).

\subsection{Results}
\label{sec:results}

\begin{figure}[t]
\centering
    \begin{minipage}[t]{0.49\textwidth}
        \centering
        {\scriptsize F1 Score}
        \includegraphics[trim = 0mm 0mm 0mm 10mm, clip, width=\textwidth]{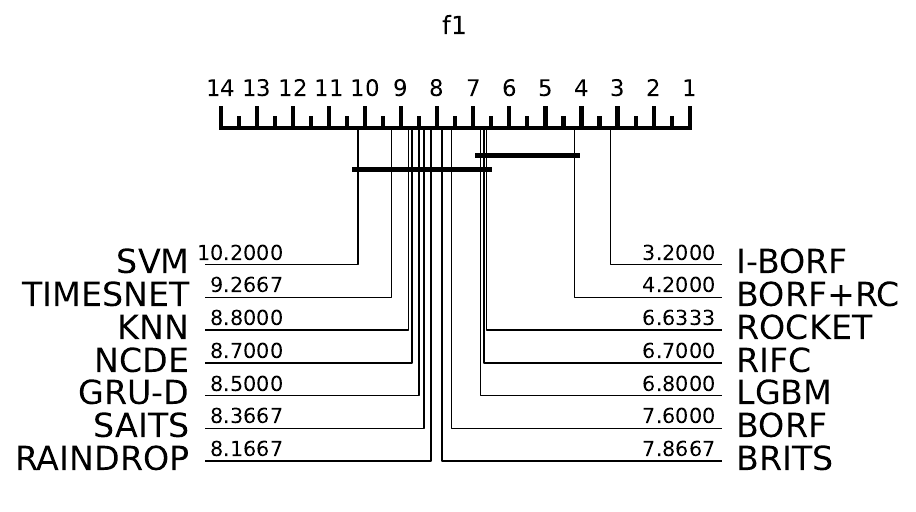}
    \end{minipage}
    \hfill
    \begin{minipage}[t]{0.49\textwidth}
        \centering
        {\scriptsize Accuracy}
        \includegraphics[trim = 0mm 0mm 0mm 10mm, clip, width=\textwidth]{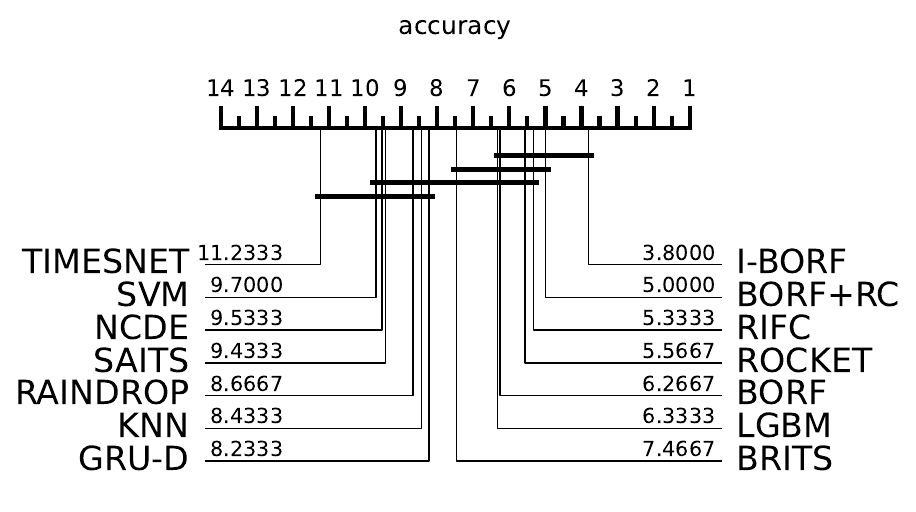}
    \end{minipage}
    \caption{CD plot for the benchmarked models in terms of F1 and Accuracy. Best models to the right. \rev{Connected methods are not significantly different according to pairwise one-sided Wilcoxon signed-rank tests with Holm correction at $\alpha=0.1$.}}
    \label{fig:cdplot}
\end{figure}

The CD plots in \Cref{fig:cdplot} provide a global comparison across datasets. 
\rev{For macro-F1, \CMethod{i-borf} is the clear leading method: it obtains the best average rank, $3.20$, and is not statistically tied to any competing approach under a pairwise one-sided Wilcoxon signed-rank test with Holm correction at $\alpha=0.1$.}
The closest method is \CMethod{borf\texttt{+}rc}, with average rank $4.20$, followed by \CMethod{rocket}, \CMethod{rifc}, and \CMethod{lgbm}. For accuracy, \CMethod{i-borf} also obtains the best average rank, $3.80$, although the separation from the closest competitors is less pronounced. \rev{Overall, the rank-based comparison indicates} indicates that the proposed transform gives the strongest aggregate performance, with an especially clear advantage in terms of macro-F1.

\begin{table}[t]
\caption{Average F1 score on the test set for each dataset and each classifier. Standard deviation is reported for highly stochastic methods. Missing values are due to exceeded memory or maximum runtime.}
\label{tab:results}
\footnotesize
\centering
\setlength{\tabcolsep}{0.9mm}
\begin{tabular}{l|c|ccccccc}
\toprule
 & \CMethod{i-borf} & \CMethod{borf} & \CMethod{borf\texttt{+}rc} & \CMethod{brits} & \CMethod{lgbm} & \CMethod{raindrop} & \CMethod{rifc} & \CMethod{rocket} \\
 \midrule
\CDatasetAbbr{ABF} & \textbf{0.90}  & 0.17  & 0.33  & 0.33 \scriptsize{$\pm$ 0.01} & 0.17  & 0.27 \scriptsize{$\pm$ 0.01} & 0.17 \scriptsize{$\pm$ 0.00} & 0.17 \scriptsize{$\pm$ 0.00} \\
\CDatasetAbbr{AN} & 0.84  & 0.80  & 0.84  & 0.65 \scriptsize{$\pm$ 0.00} & 0.80  & 0.64 \scriptsize{$\pm$ 0.05} & 0.88 \scriptsize{$\pm$ 0.02} & \textbf{0.90} \scriptsize{$\pm$ 0.04} \\
\CDatasetAbbr{DD} & \textbf{0.55}  & 0.51  & 0.52  & 0.52 \scriptsize{$\pm$ 0.02} & 0.52  & 0.45 \scriptsize{$\pm$ 0.04} & 0.49 \scriptsize{$\pm$ 0.04} & 0.54 \scriptsize{$\pm$ 0.03} \\
\CDatasetAbbr{DG} & 0.81  & 0.34  & \textbf{0.82}  & 0.72 \scriptsize{$\pm$ 0.07} & 0.34  & 0.60 \scriptsize{$\pm$ 0.22} & 0.34 \scriptsize{$\pm$ 0.00} & 0.34 \scriptsize{$\pm$ 0.00} \\
\CDatasetAbbr{DW} & \textbf{0.97}  & 0.42  & 0.96  & 0.93 \scriptsize{$\pm$ 0.02} & 0.42  & 0.78 \scriptsize{$\pm$ 0.31} & 0.42 \scriptsize{$\pm$ 0.00} & 0.42 \scriptsize{$\pm$ 0.00} \\
\CDatasetAbbr{GS} & 0.40  & \textbf{0.41}  & 0.37  & - & 0.13  & - & 0.07 \scriptsize{$\pm$ 0.02} & 0.31 \scriptsize{$\pm$ 0.15} \\
\CDatasetAbbr{LPA} & \textbf{0.96}  & 0.73  & 0.91  & 0.28 \scriptsize{$\pm$ 0.20} & 0.53  & 0.33 \scriptsize{$\pm$ 0.09} & 0.32 \scriptsize{$\pm$ 0.20} & 0.02 \scriptsize{$\pm$ 0.01} \\
\CDatasetAbbr{MI3} & 0.35  & 0.27  & 0.35  & 0.42 \scriptsize{$\pm$ 0.11} & 0.41  & 0.36 \scriptsize{$\pm$ 0.15} & \textbf{0.56} \scriptsize{$\pm$ 0.22} & 0.35 \scriptsize{$\pm$ 0.00} \\
\CDatasetAbbr{P12} & 0.57  & 0.51  & 0.56  & 0.46 \scriptsize{$\pm$ 0.00} & 0.55  & 0.56 \scriptsize{$\pm$ 0.02} & \textbf{0.63} \scriptsize{$\pm$ 0.01} & 0.47 \scriptsize{$\pm$ 0.01} \\
\CDatasetAbbr{P19} & 0.66  & 0.71  & 0.66  & 0.49 \scriptsize{$\pm$ 0.00} & \textbf{0.75}  & 0.69 \scriptsize{$\pm$ 0.01} & 0.66 \scriptsize{$\pm$ 0.03} & 0.71 \scriptsize{$\pm$ 0.01} \\
\CDatasetAbbr{PA2} & \textbf{0.75}  & 0.53  & \textbf{0.75}  & - & 0.33  & - & 0.37 \scriptsize{$\pm$ 0.32} & 0.66 \scriptsize{$\pm$ 0.10} \\
\CDatasetAbbr{PGE} & \textbf{0.78}  & 0.40  & \textbf{0.78}  & \textbf{0.78} \scriptsize{$\pm$ 0.00} & 0.40  & 0.48 \scriptsize{$\pm$ 0.26} & 0.40 \scriptsize{$\pm$ 0.00} & 0.40 \scriptsize{$\pm$ 0.00} \\
\CDatasetAbbr{SE} & 0.63  & 0.47  & 0.56  & 0.48 \scriptsize{$\pm$ 0.15} & 0.42  & 0.40 \scriptsize{$\pm$ 0.15} & \textbf{0.82} \scriptsize{$\pm$ 0.04} & 0.80 \scriptsize{$\pm$ 0.08} \\
\CDatasetAbbr{TA} & 0.44  & 0.42  & 0.44  & 0.23 \scriptsize{$\pm$ 0.00}  & \textbf{0.77}  & 0.25 \scriptsize{$\pm$ 0.02} & 0.38 \scriptsize{$\pm$ 0.06} & 0.57 \scriptsize{$\pm$ 0.01} \\
\CDatasetAbbr{VE} & 0.96  & \textbf{0.97}  & 0.92  & 0.50 \scriptsize{$\pm$ 0.08} & 0.94  & 0.65 \scriptsize{$\pm$ 0.02} & 0.90 \scriptsize{$\pm$ 0.02} & 0.94 \scriptsize{$\pm$ 0.02} \\
\bottomrule
\end{tabular}
\end{table}

The per-dataset results in \Cref{tab:results} confirm that the improvement is not driven by a single dataset. \CMethod{i-borf} obtains the best or tied-best F1 score on $6$ out of $15$ datasets, and improves over the original \CMethod{borf} on $12$ datasets. 
The comparison with \CMethod{borf\texttt{+}rc} isolates the role of the classifier. On several datasets, \CMethod{borf\texttt{+}rc} already improves substantially over the original \borf{} configuration, indicating that the linear Ridge head is a strong and stable choice for the resulting sparse representation. Nevertheless, \CMethod{i-borf} improves over \CMethod{borf\texttt{+}rc} on $8$ datasets, ties on $6$, and is worse on only one. This suggests that the gains are not only due to the classifier, but also to the proposed time-aware normalization.
Compared with competitor irregular time-series models, such as \CMethod{rifc}, \CMethod{rocket}, and \CMethod{lgbm} performance is more uniform. Indeed while the latter obtain the best score on some datasets, for instance \CDatasetAbbr{MI3}, \CDatasetAbbr{SE}, \CDatasetAbbr{TA}, and \CDatasetAbbr{P19} their performance is less uniform across the benchmark, whereas \CMethod{i-borf} provides strong average ranks.

\begin{figure}[t]
    \centering
    \includegraphics[width=.98\linewidth]{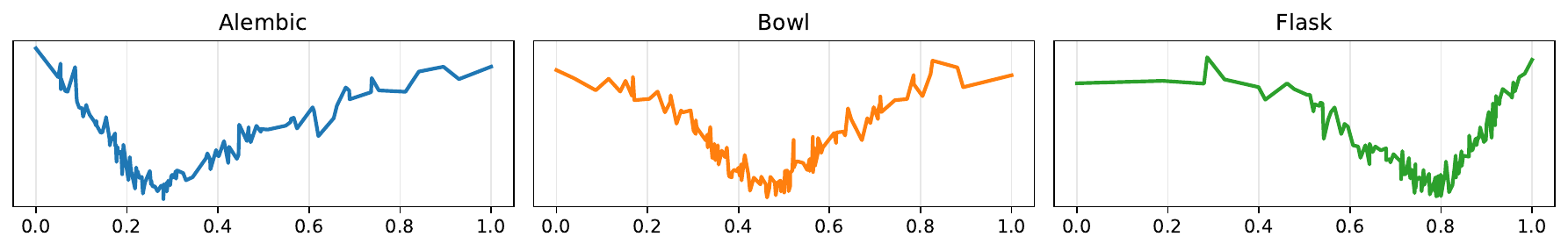}
    \caption{Three examples of instances from the \texttt{ABF} dataset, from left to right, Alembic, Bowl, and Flask.}
    \label{fig:abf}
\end{figure}

\subsection{Explainability}
\label{sec:xai}

A key advantage of \iborf{} over all other ITS classifiers considered here is the availability of human-interpretable explanations. We demonstrate this on the Alembics-Bowls-Flasks dataset (\texttt{ABF})~\cite{spinnato2026pyrregular}. There are three classes, which are Alembics, Bowls, and Flasks, and differ by how much the temporal axis is skewed, i.e., if it has positive (Alembic), negative (Flask), or no skewness (Bowl), as shown in \Cref{fig:abf}.
We compute SHAP values for a ridge classifier trained on the \iborf{}
representation and build a saliency map by projecting pattern importances
back to their temporal locations in the original signal,
following~\cite{spinnato2024fast}.

\begin{figure}[t]
\centering
    \includegraphics[trim = 0mm 0mm 0mm 0mm, clip, width=0.99\textwidth]{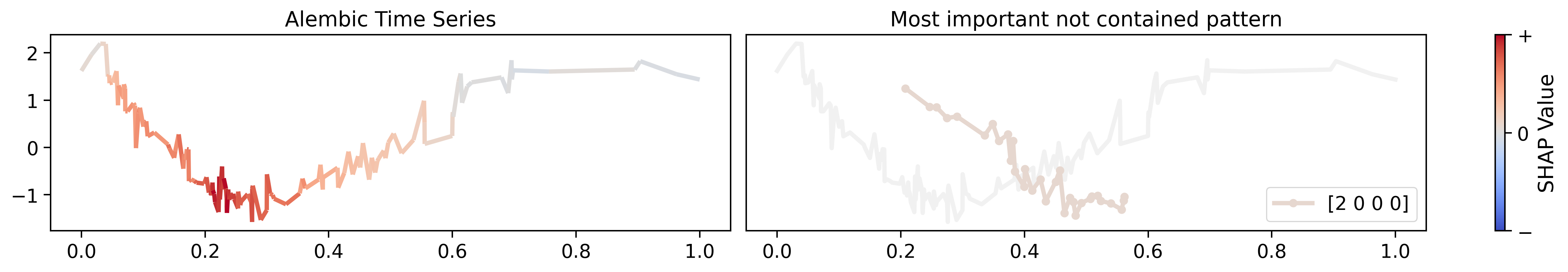}
    \includegraphics[trim = 0mm 0mm 0mm 0mm, clip, width=0.99\textwidth]{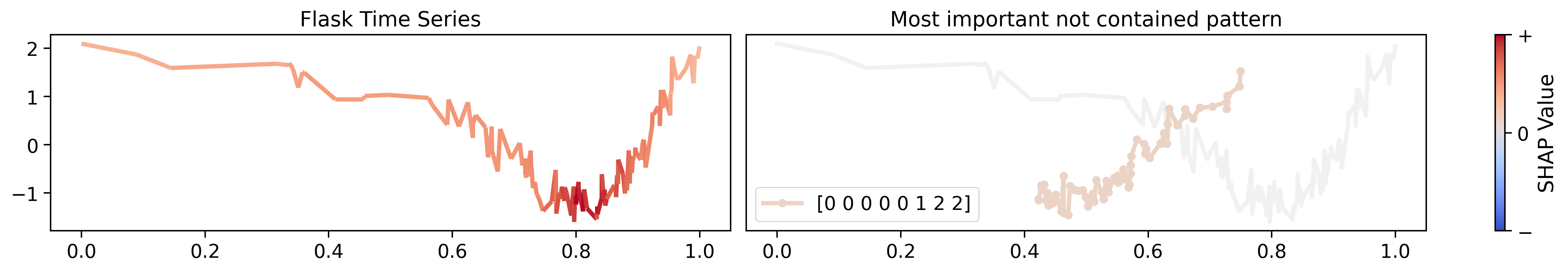}
    \caption{Local explanations on one Alembic (top) and one Flask (bottom) instance. To the left of each plot is the time series, colored based on the importance of each observation. To the right the medoid shape of the most important not contained pattern.}
    \label{fig:localexp}
\end{figure}

\medskip
\noindent
\textbf{Local explanation.}
\Cref{fig:localexp} shows local explanations for one Alembic instance and one Flask instance. In both cases, the left panel reports the original time series, with observations colored according to their SHAP contribution, while the right panel shows the medoid of the most important pattern that is not contained in the instance. For the Alembic instance, the strongest contribution is concentrated in the early-to-middle part of the series, where the signal follows a fast decreasing trajectory before reaching its minimum. The most important not-contained pattern is also decreasing, but slowly, more characteristic of the opposite skew (associated with Flasks). For the Flask instance, the relevant region is shifted toward the later part of the series, where the curve rises after a pronounced valley. In this case, the most important not-contained pattern is slowly increasing, closer to Alembic instances.

\begin{figure}[t]
\centering
    \includegraphics[trim = 0mm 0mm 0mm 0mm, clip, width=0.99\textwidth]{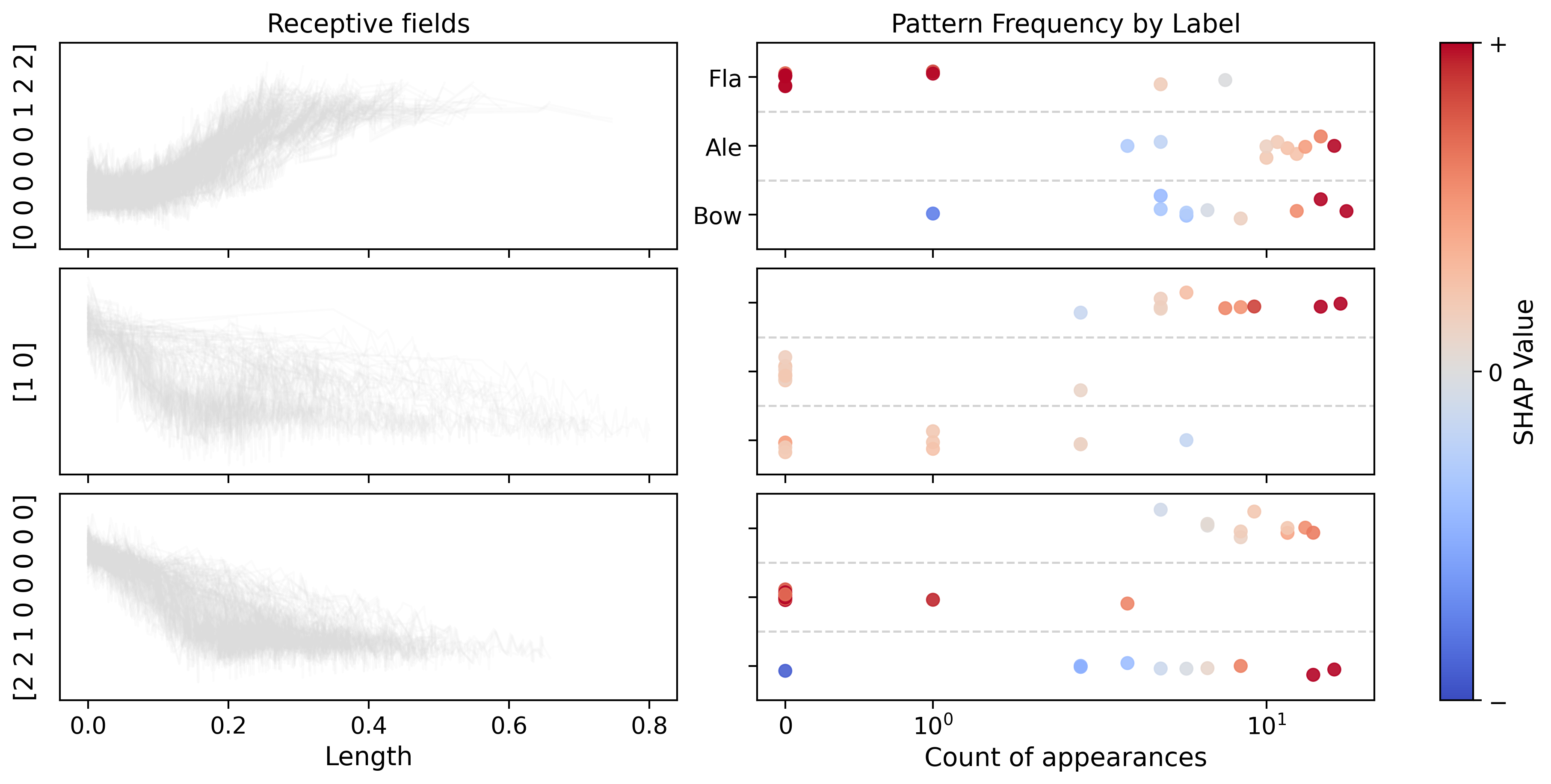}
    \caption{Global explanations on the \texttt{ABF} dataset. Left: all temporal alignments of the receptive fields associated with the most important symbolic patterns. Right: pattern frequency by class, with points colored according to their SHAP value.}
    \label{fig:globalexp}
\end{figure}

\medskip
\noindent
\textbf{Global explanation.}
\Cref{fig:globalexp} provides a dataset-level view of the top-3 most important patterns. The left panels show the receptive fields associated with the selected symbolic patterns, in all their possible alignments in the training set, while the right panels report how often each pattern appears in the three classes, with color indicating its SHAP value. The visualization makes the class structure explicit. Patterns associated with Alembics tend to correspond to slowly decreasing receptive fields, while patterns associated with Flasks exhibit the opposite behavior. Bowls occupy an intermediate regime: their relevant patterns are less skewed and appear between the two extremes represented by Alembics and Flasks.
The frequency plots further show that the learned representation is not only separating the classes through raw occurrence counts, but also through the direction and magnitude of the induced classifier contribution. Patterns that occur frequently in one class and receive consistently positive SHAP values provide global evidence for that class, whereas absent or negatively weighted patterns provide evidence against it. This illustrates that \iborf{} retains the interpretability mechanism of \borf{} while extending it to irregular time series: explanations can still be read both locally, as salient temporal regions in a single instance, and globally, as class-specific receptive-field shapes.

% -----------------------------------------------------------------------
\section{Conclusions}
\label{sec:conclusions}

\rev{We presented \iborf{}, an extension of \borf{} to irregular time series that introduces time-weighted aggregation and normalization into the symbolic transformation.} This makes the transform aware of the temporal distribution of samples without imputation, while preserving the linear complexity, determinism, and interpretability of \borf{}. Its main technical contribution is a sliding-window weighted standard deviation recurrence, which extends the moving-average speedup of \borf{} and Welford-style updates to weighted sliding windows. Experiments on the \texttt{pyrregular} benchmark show that time-weighting improves over the unweighted baseline. Moreover, \iborf{} retains the explanation framework of \borf{}, supporting local saliency maps and global pattern-importance analyses, a capability absent from the competing ITS classifiers considered here.

\rev{A limitation is that \iborf{} introduces time awareness only in segment and window statistics, while receptive-field construction and word counts remain index-based. Moreover, time-delta weighting assumes that each observation represents the preceding interval, which may be inappropriate for point events or clinician-triggered measurements, and does not explicitly model partial observation. Future work will study alternative weighting conventions, the effects of different irregularity types, continuous-time receptive fields, and the quantitative quality and stability of explanations.}

\paragraph{Acknowledgements.}
This study has been partially funded by the Italian Project Fondo Italiano per la Scienza FIS00001966 ``MIMOSA'', 101120763 ``TANGO'', by the European Commission under the NextGeneration EU programme, ``SoBigData.it – Strengthening the Italian RI for Social Mining and Big Data Analytics'' – Prot. IR0000013 –  Av. n. 3264 del 28/12/2021.

% -----------------------------------------------------------------------
\bibliographystyle{splncs04}
\bibliography{biblio}

% -----------------------------------------------------------------------
\appendix

\section{Proof of Proposition~\ref{prop:wvar}}
\label{sec:derivation}

Throughout, $W_i = \sum_{r=i}^{i+w-1} \delta_r$ denotes the total
weight of the window $[x_i, \ldots, x_{i+w-1}]$, and
$\hat{\mu}_i = (1/W_i)\sum_{r=i}^{i+w-1} \delta_r\, x_r$
its weighted mean.
We write $W_{i+1} = W_i - \delta_i + \delta_{i+w}$.

\paragraph{Proof.}
Define $S_i = W_i\,(\hat{\sigma}_i)^2$.
Using the identity
\begin{equation}
  (\hat{\sigma}_i)^2
  = \frac{1}{W_i}\sum_{r=i}^{i+w-1} \delta_r( x_r - \hat{\mu}_i)^2
 = \frac{1}{W_i}\left[\sum_{r=i}^{i+w-1} \delta_r\, x_r^2\right] - \hat{\mu}_i^2,
\end{equation}

we obtain:
\begin{equation}
  S_i
  = \left[\sum_{r=i}^{i+w-1} \delta_r\, x_r^2\right]
    - W_i\,\hat{\mu}_i^2.
  \label{eq:app:S}
\end{equation}
Computing $S_{i+1} - S_i$ and substituting \Cref{eq:app:S}:
\begin{align}
  S_{i+1} - S_i
    &= \delta_{i+w}\, x_{i+w}^2 - \delta_i\, x_i^2
       - W_{i+1}\,\hat{\mu}_{i+1}^2
       + W_i\,\hat{\mu}_i^2.
  \label{eq:app:Sdiff}
\end{align}
Substituting $W_i = W_{i+1} - \delta_{i+w} + \delta_i$
into the last term and grouping:
\begin{align}
  &- W_{i+1}\,\hat{\mu}_{i+1}^2 + W_i\,\hat{\mu}_i^2
  \notag \\
  &= W_{i+1}(\hat{\mu}_i - \hat{\mu}_{i+1})
                 (\hat{\mu}_i + \hat{\mu}_{i+1})
     + \delta_i\,\hat{\mu}_i^2
     - \delta_{i+w}\,\hat{\mu}_i^2.
  \label{eq:app:expand}
\end{align}
Multiplying $\hat{\mu}_{i+1}$ (\Cref{eq:wmean_update}) through by $W_{i+1}$ and rearranging,
using $W_{i+1} - W_i = \delta_{i+w} - \delta_i$:
\begin{align}
  W_{i+1}(\hat{\mu}_i - \hat{\mu}_{i+1})
    &= W_{i+1}\hat{\mu}_i - W_i\hat{\mu}_i
       + \delta_i x_i - \delta_{i+w} x_{i+w} \notag \\
    &= (\delta_{i+w} - \delta_i)\hat{\mu}_i
       + \delta_i x_i - \delta_{i+w} x_{i+w} \notag \\
    &= \delta_i(x_i - \hat{\mu}_i)
       - \delta_{i+w}(x_{i+w} - \hat{\mu}_i).
  \label{eq:app:diff}
\end{align}
Substituting \Cref{eq:app:diff} into \Cref{eq:app:expand} and then into
\Cref{eq:app:Sdiff}:
\begin{align}
  S_{i+1} - S_i
    &= \delta_{i+w}\, x_{i+w}^2 - \delta_i\, x_i^2 \notag \\
    &\quad + \bigl[\delta_i(x_i - \hat{\mu}_i)
                 - \delta_{i+w}(x_{i+w} - \hat{\mu}_i)\bigr]
             (\hat{\mu}_i + \hat{\mu}_{i+1}) \notag \\
    &\quad + \delta_i\,\hat{\mu}_i^2 - \delta_{i+w}\,\hat{\mu}_i^2.
\end{align}
Collecting terms in $\delta_{i+w}$ and $\delta_i$:
\begin{align}
  S_{i+1} - S_i
    &= \delta_{i+w}\bigl[
         x_{i+w}^2
         - (x_{i+w} - \hat{\mu}_i)(\hat{\mu}_i + \hat{\mu}_{i+1})
         - \hat{\mu}_i^2
       \bigr] \notag \\
    &\quad + \delta_i\bigl[
         -x_i^2
         + (x_i - \hat{\mu}_i)(\hat{\mu}_i + \hat{\mu}_{i+1})
         + \hat{\mu}_i^2
       \bigr].
\end{align}

Using the algebraic identities
\[
x^2 - (x-\mu_{\mathrm{old}})(\mu_{\mathrm{old}}+\mu_{\mathrm{new}})
 - \mu_{\mathrm{old}}^2
 = (x-\mu_{\mathrm{old}})(x-\mu_{\mathrm{new}})
\]
and
\[
-x^2 + (x-\mu_{\mathrm{old}})(\mu_{\mathrm{old}}+\mu_{\mathrm{new}})
 + \mu_{\mathrm{old}}^2
 = -(x-\mu_{\mathrm{old}})(x-\mu_{\mathrm{new}}),
\]
for the entering and leaving observations, respectively, gives
\Cref{eq:wvar_update}. \qed

\end{document}